\documentclass[10pt,twocolumn,letterpaper]{article}

\usepackage{wacv}

\usepackage{graphicx}
\usepackage{amsmath}
\usepackage{amssymb}
\usepackage{booktabs}

\usepackage[pagebackref,breaklinks,colorlinks]{hyperref}

\usepackage[capitalize]{cleveref}
\crefname{section}{Sec.}{Secs.}
\Crefname{section}{Section}{Sections}
\Crefname{table}{Table}{Tables}
\crefname{table}{Tab.}{Tabs.}

\def\wacvPaperID{*****} 
\def\confName{WACV}
\def\confYear{2025}

\begin{document}

\title{Label Errors in the Tobacco3482 Dataset}

\author{Gordon Lim\\
University of Michigan\\
Ann Arbor, MI, USA\\
{\tt\small gbtc@umich.edu}
\and
Stefan Larson\\
Vanderbilt University\\
Nashville, TN, USA\\
{\tt\small stefan.larson@vanderbilt.edu}
\and
Kevin Leach\\
Vanderbilt University\\
Nashville, TN, USA\\
{\tt\small kevin.leach@vanderbilt.edu}
}
\maketitle

\begin{abstract}
Tobacco3482 is a widely used document classification benchmark dataset. 
However, our manual inspection of the entire dataset uncovers widespread ontological issues, especially large amounts of annotation label problems in the dataset.
We establish data label guidelines and find that 11.7\% of the dataset is improperly annotated and should either have an unknown label or a corrected label, and 16.7\% of samples in the dataset have multiple valid labels.
We then analyze the mistakes of a top-performing model and find that 35\% of the model's mistakes can be directly attributed to these label issues, highlighting the inherent problems with using a noisily labeled dataset as a benchmark.
Supplementary material, including dataset annotations and code, is available at \href{https://github.com/gordon-lim/tobacco3482-mistakes/}{github.com/gordon-lim/tobacco3482-mistakes/}.
\end{abstract}

\section{Introduction}
Document classification is a fundamental component of document processing pipelines used by organizations for efficient search and retrieval~\cite{rvlcdip}.
Recent advances in document classification have demonstrated continuous improvement in performance on the Tobacco3482 \cite{LogoDetection-ICDAR07}  document image dataset (e.g., ~\cite{docxclassifier-saifullah2022,squeezenet-hassanpour2019,asim2019two,noce2016embedded}).
However, a growing body of research on dataset quality casts serious doubt on the usefulness of many benchmark datasets for evaluating model performance~\cite{laroca2023-duplicates, rvlcdip-problems, ying-thomas-2022-label, jimaging6060041}. 
Notably, widely-used datasets like ImageNet~\cite{imagenet} has been found to have labeling issues, including incorrect labels and images that should be assigned multiple labels~\cite{northcutt2021confident, northcutt2021pervasive, luccioni2024the-nine-lives-imagenet}.
Such issues destabilize benchmarks and make low capacity models seem less performant than they are~\cite{northcutt2021pervasive}.
To our knowledge, no studies have yet quantified the extent of label issues in the Tobacco3482 dataset.

\begin{figure}[t]
    \centering
    \includegraphics[width=\linewidth]{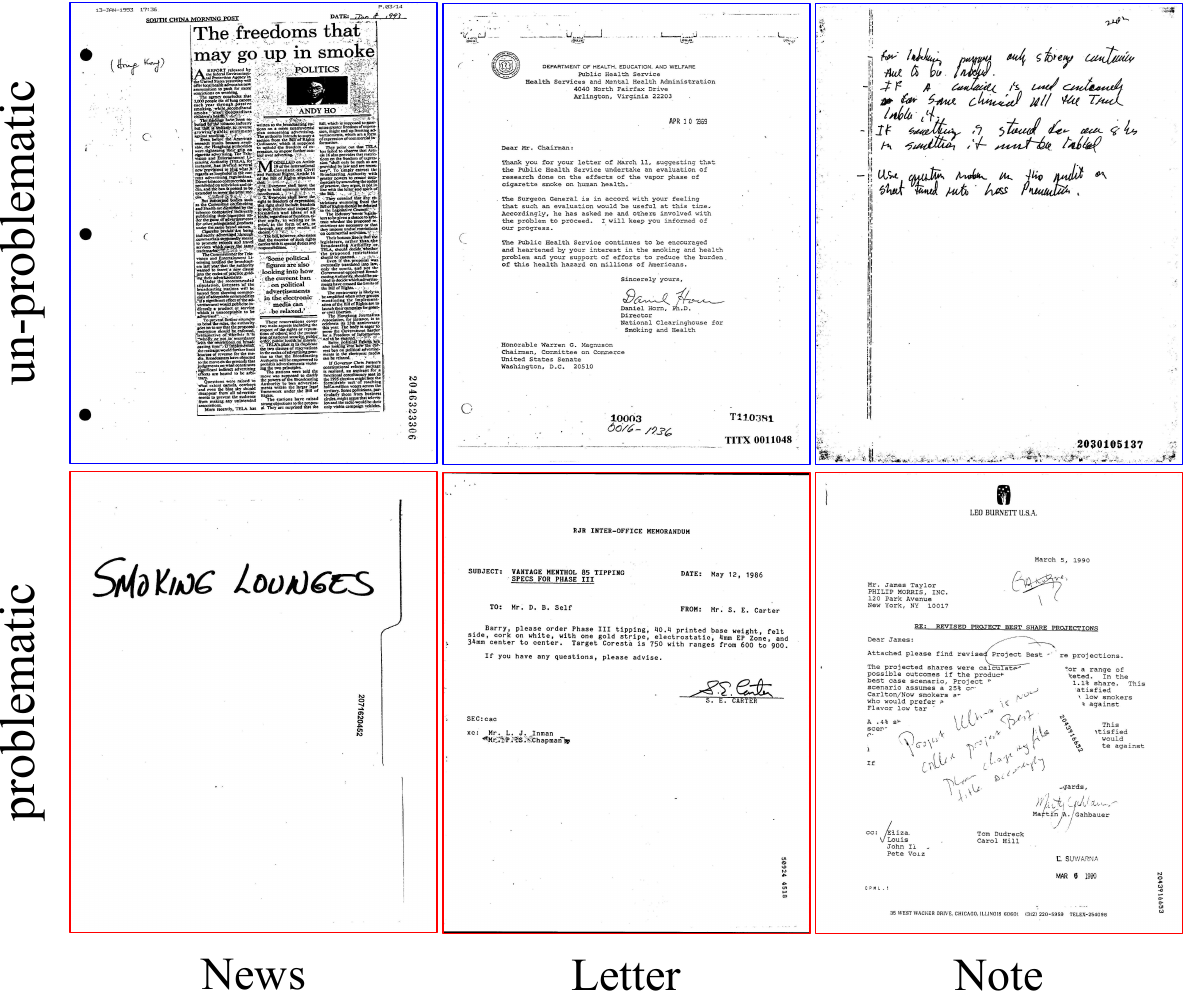}
    \caption{Top row: un-problematic document images from Tobacco3482. Bottom row: samples from Tobacco3482 that are erroneously labeled (left and center) or could have multiple valid labels (right). Bottom left: the document is not a news article. Bottom center: the document should be labeled as Memo and not Letter. Bottom right: the image contains both a letter (background) and a note (foreground), and thus could have two valid labels (Note and Letter).}
    \label{fig:teaser}
\end{figure}

This short paper presents the first examination of the Tobacco3482 dataset for data label issues. 
We conduct a thorough review of the Tobacco3482 dataset, and find that 11.7\% of samples from the dataset are either mis-labeled or do not belong to any of Tobacco3482's labels, and 16.7\% of samples from the dataset have multiple valid labels.
Examples of data labeling issues are shown in Figure~\ref{fig:teaser}.
To contextualize the impact of these data labeling issues on model evaluation, we analyze the mistakes of a top-performing transformer-based document image classification model and find that 35\% of the model's mistakes are actually valid alternative labels or images that were not represented by any of the dataset's labels. 
Overall, our findings highlight flaws in the Tobacco3482 dataset concerining data quality.

\section{A Review of Tobacco3482}

\begin{table*}[t]
\centering
    \scalebox{0.7}{
    \begin{tabular}{l p{18cm} }
        \hline
        \textbf{Category} & \textbf{Label Guidelines}\\
        \hline
        \texttt{ADVE} & Media targeted towards a general audience to promote a product, service, or agenda. Can be found in newspapers. Excludes magazine covers and content drafts. \\
        \hline
        \texttt{Email} & Electronic mail, often including ``Sent via electronic mail'' or timestamps. Common features are ``To/From'', ``Subject'', ``cc'' fields, and digital attachments. Excludes HTML code. \\
        \hline
        \texttt{Form} & Documents with spaces (boxes, lines, etc.) to be filled in. Examples include registration forms, report templates, and data sheets. Excludes documents that print ``Form" but do have fillable spaces. \\
        \hline
        \texttt{Letter} & Messages delivered on physical paper with physical addresses. Features include a letterhead, recipient’s name, address, salutation (e.g., ``Dear xxx''), and sign-off (e.g., ``Sincerely,''). Excludes \textit{memos} using ``letter'' in its subject. \\
        \hline
        \texttt{Memo} & Documents addressed to an organization or person, often denoted by ``To/From”. They might state ``Memo'', ``Memorandum'', or ``Correspondence''. Typical content include policy changes, meeting schedules, event invitations, recommendations, or actions. Excludes documents that print ``File note'' not addressed to anyone. \\
        \hline
        \texttt{News} & Articles resembling those in newspapers, typically with multiple columns and select font enlarged. They may also include newspaper clippings. Excludes scientific articles published in scientific news platforms. \\
        \hline
        \texttt{Note} & Handwritten or typed brief messages. Typically smaller than letter-sized paper and have minimal layout features. Excludes handwriting on other documents and \textit{Reports} with ``Note'' in title. \\
        \hline
        \texttt{Report} & Formal documents presenting information, often with section headers such as introduction, methodology, and results. Topics can be scientific or non-scientific and might include data reports, findings, or series of events. May also presented using bullet lists or tables. Excludes documents that are primarily applications, requests, or proposals. \\
        \hline
        \texttt{Resume} & Formal documents containing personal career and/or education information, often titled ``Curriculum Vitae'', ``Resume'' or ``Biographical Sketch''. They include personal information like name, title, birth date, education, and professional experience. \\
        \hline
        \texttt{Scientific} & Documents containing scientific knowledge, typically written by scientists, often featuring scientific terms, mathematical formulas, and section headers like Abstract. Examples include research papers, grant applications, and project reports. Excludes business reports and product development reports. \\
        \hline \\
    \end{tabular}
    } 
    \caption{Guidelines and examples illustrating how we assigned new labels for the Tobacco3482 dataset.}
    \label{tab:label-guidelines}
\end{table*}

In this section we first discuss the background and content of Tobacco3482, then we conduct a review of the dataset in order to establish label guidelines.

\subsection{Background on Tobacco3482}

Tobacco3482 \cite{LogoDetection-ICDAR07} consists of 3,482 document images, each labeled as one of 10 categories:
Advertisement (ADVE), Email, Form, Letter, Memo, News, Note, Report, Resume, or Scientific.
These images were sampled from the IIT Complex Document Information Processing (IIT-CDIP) Test Collection, which in turn sources its documents from the Truth Tobacco Industry Documents (TTID).\footnote{Available at: \href{https://www.industrydocuments.ucsf.edu/tobacco/}{industrydocuments.ucsf.edu/tobacco/} TTID was previously known as the Legacy Tobacco Documents Library (LTDL).}
TTID is a document collection containing internal documents from major US tobacco companies, released as part of legal settlements~\cite{iit-cdip}. 
Documents in TTID have one or multiple manual document category annotations\footnote{Larson, et al. (2023) \cite{rvlcdip-problems} hypothesize that these category labels were assigned by human annotators employed by the TTID's parent organization (the LTDL).} that include the 10 Tobacco3482 categories. 
Using TTID’s API, we searched for Tobacco3482 documents by searching for the Bates numbers in their filenames, resulting in the identification of 1,707 documents.
Of these, 134 documents (3.8\% of the entire Tobacco3482 dataset) have multiple Tobacco3482 category annotations.
This indicates that at least some of the documents from Tobacco3482 may have valid alternate labels.
Despite this, the Tobacco3482 dataset still assigns only a single label to all of its documents.
Furthermore, there are no formal labeling guidelines reported by Tobacco3482, IIT-CDIP, or TTID, which raises concerns about the consistency and reliability of the labels within the Tobacco3482 dataset.


\begin{figure*}
    \centering
    \includegraphics[width=0.85\linewidth]{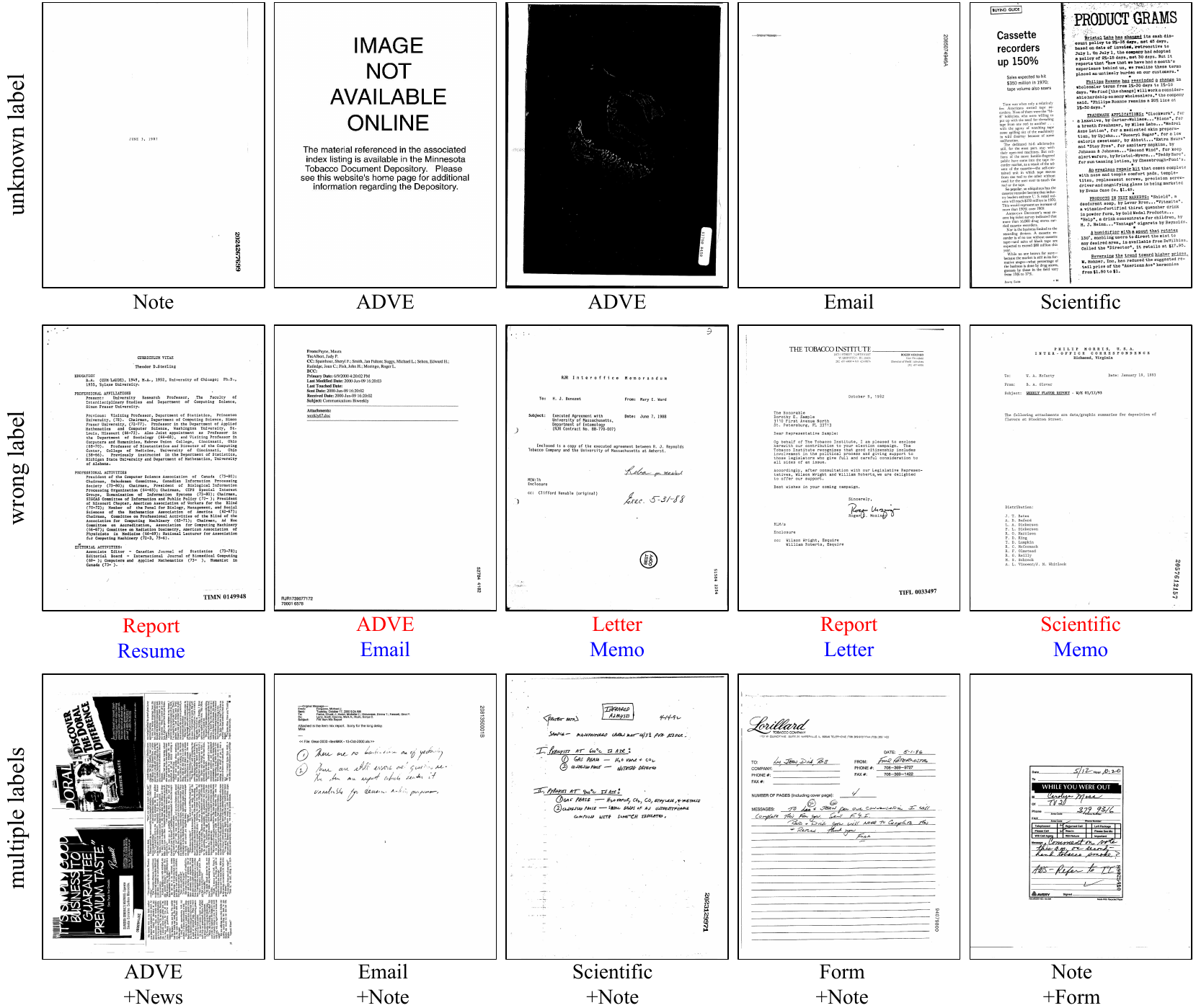}
    \caption{Examples of problematic samples from Tobacco3482. Top row: documents where the valid label is unknown. Middle row: documents that have the wrong original label (shown in red (top label) with corrected label shown in blue (bottom label)). Bottom row: documents that have multiple valid labels (original label shown on top, with additional valid label shown on bottom).}
    \label{fig:examples_errors_3x5}
\end{figure*}

\subsection{Data Label Guidelines and Review}

We seek to quantify the amount of label errors in the Tobacco3482 dataset.
Since the Tobacco3482 dataset (and its parent TTID corpus) contains no labeling guidelines, we conducted an initial review of the Tobacco3482 dataset in order to establish annotation guidelines.
We documented the process and provide a brief version of our guidelines, along with descriptive examples, in Table~\ref{tab:label-guidelines}.
This procedure was also conducted in prior work by Larson, et al. (2023) \cite{rvlcdip-problems}, who established label guidelines in order to analyze label error rates in the RVL-CDIP dataset.

Like Larson, et al. (2023) \cite{rvlcdip-problems}, we then reviewed the Tobacco3482 dataset using our guidelines, and re-annotated any document images that were problematically labeled.
We tracked three types of problematic label types: (1) \emph{unknown}, where the document image does not fit within the guidelines of any of the 10 Tobacco3482 categories; (2) \emph{mis-labeled}, where the document image should be re-annotated with a different, more correct Tobacco3482 category; and (3) \emph{multiple labels}, where the document image contains multiple valid category labels.
Examples of each problematic label type are displayed in Figure~\ref{fig:examples_errors_3x5}.
We quantify the amount of each problematic label type below.

\paragraph{Unknown and Wrong Labels.}
We found 151 documents that did not belong to any of the 10 Tobacco3482 categories according to our guidelines in Table~\ref{tab:label-guidelines}.
Additionally, we found 258 documents that were incorrectly labeled and have a valid alternative.
Together, these two label issues account for 409 samples, or 11.7\% of the entire Tobacco3482 dataset.
Table~\ref{tab:counts} displays the rates of each label error type for each Tobacco3482 category.
The Scientific category contains many errors; roughly 14.6\% of its documents  are unknown, and 23.8\% documents are mis-labeled.
Roughly 24\% of the Letter category is mis-labeled, where most of the corrections should be to the Memo category (see the companion repository for a confusion matrix\footnote{\href{https://github.com/gordon-lim/tobacco3482-mistakes/}{github.com/gordon-lim/tobacco3482-mistakes/}}).

\begin{figure*}[th]
    \centering
    \includegraphics[width=0.75\linewidth]{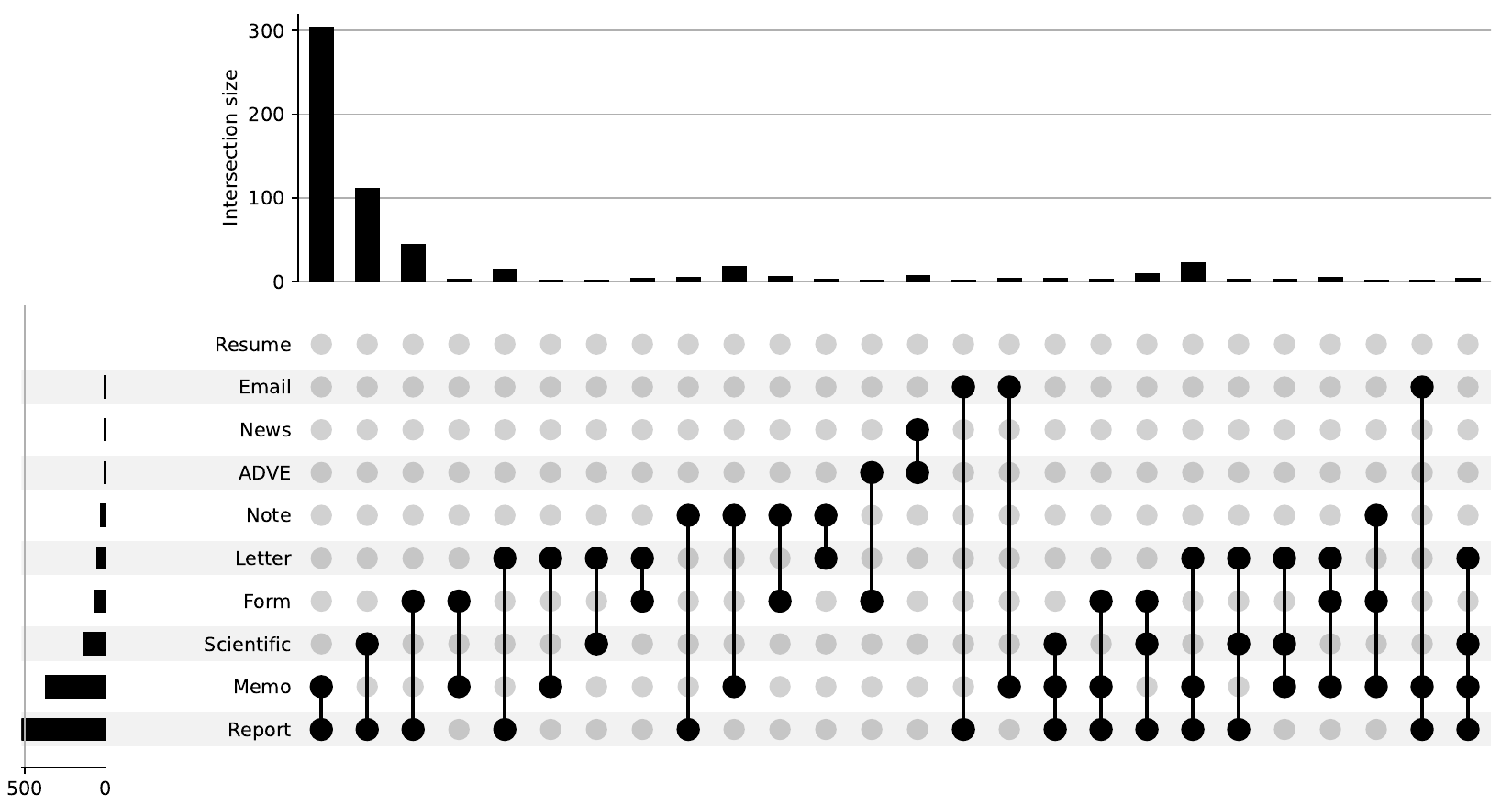}
    \caption{UpSet plot~\cite{upset-plot-2014} of multi-label Tobacco3482 category annotations. A majority of the documents with multiple labels are documents that are both Reports and Memos.}
    \label{fig:multi-plot}
\end{figure*}

\paragraph{Multiple Lables.}
We identified 583 samples (16.7\% of the dataset) that should have multiple valid Tobacco3482 category labels.
Figure~\ref{fig:multi-plot} compares the overlap between different Tobacco3482 categories on our multi-label annotations.
The Memo and Report categories exhibit the most overlap, as reports can often be written in memo format for communication purposes. 
The Report and Scientific categories also contain much overlap, since report-style documents can encapsulate the results of scientific studies, and vice-versa.
(See the companion repository for examples.)

\begin{table}[]
    \centering\scalebox{0.9}{
    \begin{tabular}{lccc}
    \toprule
    \textbf{Category} & \textbf{Num. Samples}  & \textbf{$\%$ Unknown}  & \textbf{$\%$ Mis-Label}  \\
    \midrule
    ADVE & 230 & 7.39 & 2.17 \\
    Email & 598 & 2.51 & 2.68 \\
    Form & 431 & 9.05 & 4.18 \\
    Letter & 567 & 0.35 & 24.0 \\
    Memo & 620 & 0.16 & 0.48 \\
    News & 188 & 3.19 & 0.00 \\
    Note & 201 & 6.97 & 4.48 \\
    Report & 265 & 7.17 & 3.40 \\
    Resume & 119 & 0.00 & 0.00 \\
    Scientific & 261 & 14.6 & 23.8 \\
    \bottomrule
    \end{tabular}}
    \caption{Number of samples for each category, and label error rates for unknown and mis-labeled samples in Tobacco3482.}
    \label{tab:counts}
\end{table}


\section{Impact of Label Issues on the Classification Task}

To contextualize the impact of these data labeling issues on model evaluation, we analyze the mistakes of a DiT model \cite{li2022dit}.
We used a DiT model from Hugging Face\footnote{\href{https://huggingface.co/docs/transformers/en/model\_doc/dit}{huggingface.co/docs/transformers/en/model\_doc/dit}}, using RVL-CDIP~\cite{rvlcdip} pre-trained weights. 
We selected this model because it is a top-performing model on RVL-CDIP, has not yet been benchmarked on Tobacco3482, and has accessible source code.
To collect DiT's mistakes on the entire original (i.e., un-corrected) Tobacco3482 dataset, we performed our evaluation using 4-fold cross-validation, achieving a 84.1\% top-1 accuracy.


Following the above, we found that of the 554 mistakes made by the DiT model on the un-corrected Tobacco3482 dataset, 196 of them were not actually mistakes. 
Specifically, in 147 instances, the model predicted an alternative valid label from our multi-label annotations. 
Additionally, in 49 instances, the model faced the unfair and impossible challenge of predicting an unknown label.
If we consider these \textit{mistakes} as correct, the DiT model's accuracy would increase to 89.7\%. This marks a 5.7\% improvement from the original 84.0\% accuracy and brings it closer to a more recent method that achieved 90.7\% accuracy using four times more parameters~\cite{docxclassifier-saifullah2022}.

\section{Discussion}
We inspected the Tobacco3482 dataset for problematic annotation labels and found there to be a substantial amount of various types of label issues.
Within the document understanding realm, this finding is unfortunately shared in common with the larger RVL-CDIP document classification corpus, in which Larson, et al. (2023) \cite{rvlcdip-problems} observed similar label problems.
Thus, published model performance on the original versions of Tobacco3482 and RVL-CDIP may not be indicative of a model's capabilities.
In a recent orthogonal finding to ours, Saifullah, et al. (2024) \cite{saifullah-rvl-tobacco-reality-2024} observed that both Tobacco3482 and RVL-CDIP contain significant feature biases like numeric codes in the document images that impact model performance.
Like us, Saifullah, et al. raise concern over whether published model performance on the original Tobacco3482 dataset is a valid measure of a model's true capabilities.
Moreover, recent work by Larson, et al. (2024) \cite{larson-etal-2024-de} found large amounts of sensitive personal information in both Tobacco3482 and RVL-CDIP; we thus caution researchers in the document understanding community on using Tobacco3482 and other datasets derived from TTID due to the presence of label errors, data bias, and sensitive material.

\section{Conclusion}

This short paper represents a first effort to measure the extent of label errors in the Tobacco3482 dataset, adding to the relevant literature alongside papers that do the same for other datasets, such as RVL-CDIP~\cite{rvlcdip-problems}.
We also present early evidence showing how such errors can make model accuracy scores on these benchmarks misleading, as many classification mistakes committed by the DiT classifier were not actually mistakes.
As model performance approaches near-perfect levels ($>$95\%), it is crucial that these accuracy metrics reflect a model's ability to learn meaningful features seen in the real world, rather than spurious features in the dataset.


{\small
\bibliographystyle{ieee_fullname}
\bibliography{egbib}

\begin{thebibliography}{10}\itemsep=-1pt

\bibitem{asim2019two}
Muhammad~Nabeel Asim, Muhammad Usman~Ghani Khan, Muhammad~Imran Malik, Khizar Razzaque, Andreas Dengel, and Sheraz Ahmed.
\newblock Two stream deep network for document image classification.
\newblock In {\em Proceedings of the 2019 International Conference on Document Analysis and Recognition (ICDAR)}, 2019.

\bibitem{jimaging6060041}
Bj{\"o}rn Barz and Joachim Denzler.
\newblock Do we train on test data? purging {CIFAR} of near-duplicates.
\newblock {\em Journal of Imaging}, 6(6), 2020.

\bibitem{imagenet}
Jia Deng, Wei Dong, Richard Socher, Li-Jia Li, Kai Li, and Li Fei-Fei.
\newblock Image{N}et: A large-scale hierarchical image database.
\newblock In {\em Proceedings of the 2009 IEEE Conference on Computer Vision and Pattern Recognition}, 2009.

\bibitem{rvlcdip}
Adam~W. Harley, Alex Ufkes, and Konstantinos~G. Derpanis.
\newblock Evaluation of deep convolutional nets for document image classification and retrieval.
\newblock In {\em Proceedings of the International Conference on Document Analysis and Recognition ({ICDAR})}, 2015.

\bibitem{squeezenet-hassanpour2019}
Mohammad Hassanpour and Hamed Malek.
\newblock Document image classification using {S}queeze{N}et convolutional neural network.
\newblock In {\em Proceedings of the 5th Iranian Conference on Signal Processing and Intelligent Systems (ICSPIS)}, 2019.

\bibitem{laroca2023-duplicates}
Rayson Laroca, Valter Estevam, Alceu~S. Britto, Rodrigo Minetto, and David Menotti.
\newblock Do we train on test data? the impact of near-duplicates on license plate recognition.
\newblock In {\em Proceedings of the 2023 International Joint Conference on Neural Networks (IJCNN)}, 2023.

\bibitem{rvlcdip-problems}
Stefan Larson, Gordon Lim, and Kevin Leach.
\newblock On evaluation of document classification with {RVL}-{CDIP}.
\newblock In {\em Proceedings of the 17th Conference of the European Chapter of the Association for Computational Linguistics (EACL)}, 2023.

\bibitem{larson-etal-2024-de}
Stefan Larson, Nicole~Cornehl Lima, Santiago~Pedroza Diaz, Amogh~Manoj Joshi, Siddharth Betala, Jamiu~Tunde Suleiman, Yash Mathur, Kaushal~Kumar Prajapati, Ramla Alakraa, Junjie Shen, Temi Okotore, and Kevin Leach.
\newblock De-identification of sensitive personal data in datasets derived from {IIT}-{CDIP}.
\newblock In {\em Proceedings of the 2024 Conference on Empirical Methods in Natural Language Processing (EMNLP)}, 2024.

\bibitem{iit-cdip}
David Lewis, Gady Agam, Shlomo Argamon, Ophir Frieder, D. Grossman, and Jefferson Heard.
\newblock Building a test collection for complex document information processing.
\newblock In {\em Proceedings of the 29th Annual International ACM SIGIR Conference on Research and Development in Information Retrieval}, 2006.

\bibitem{upset-plot-2014}
Alexander Lex, Nils Gehlenborg, Hendrik Strobelt, Romain Vuillemot, and Hanspeter Pfister.
\newblock Up{S}et: Visualization of intersecting sets.
\newblock {\em IEEE Transactions on Visualization and Computer Graphics}, 20(12):1983--1992, 2014.

\bibitem{li2022dit}
Junlong Li, Yiheng Xu, Tengchao Lv, Lei Cui, Cha Zhang, and Furu Wei.
\newblock Di{T}: Self-supervised pre-training for document image transformer.
\newblock In {\em Proceedings of the 30th ACM International Conference on Multimedia}, 2022.

\bibitem{luccioni2024the-nine-lives-imagenet}
Sasha Luccioni and Kate Crawford.
\newblock The nine lives of {I}mage{N}et: A sociotechnical retrospective of a foundation dataset and the limits of automated essentialism.
\newblock {\em Journal of Data-centric Machine Learning Research}, 2024.

\bibitem{noce2016embedded}
Lucia Noce, Ignazio Gallo, Alessandro Zamberletti, and Alessandro Calefati.
\newblock Embedded textual content for document image classification with convolutional neural networks.
\newblock In {\em Proceedings of the 2016 ACM Symposium on Document Engineering}, 2016.

\bibitem{northcutt2021confident}
Curtis Northcutt, Lu Jiang, and Isaac Chuang.
\newblock Confident learning: Estimating uncertainty in dataset labels.
\newblock {\em J. Artif. Int. Res.}, 70:1373–1411, May 2021.

\bibitem{northcutt2021pervasive}
Curtis~G. Northcutt, Anish Athalye, and Jonas Mueller.
\newblock Pervasive label errors in test sets destabilize machine learning benchmarks.
\newblock In {\em Proceedings of the 35th Conference on Neural Information Processing Systems Track on Datasets and Benchmarks}, 2021.

\bibitem{docxclassifier-saifullah2022}
Saifullah, Stefan Agne, Andreas Dengel, and Sheraz Ahmed.
\newblock Doc{X}classifier: towards a robust and interpretable deep neural network for document image classification.
\newblock {\em International Journal on Document Analysis and Recognition (IJDAR)}, 27(3), 2024.

\bibitem{saifullah-rvl-tobacco-reality-2024}
Saifullah, Stefan Agne, Andreas Dengel, and Sheraz Ahmed.
\newblock The reality of high performing deep learning models: A case study on document image classification.
\newblock {\em IEEE Access}, 12:103537--103564, 2024.

\bibitem{ying-thomas-2022-label}
Cecilia Ying and Stephen Thomas.
\newblock Label errors in {BANKING}77.
\newblock In {\em Proceedings of the Third Workshop on Insights from Negative Results in NLP}, 2022.

\bibitem{LogoDetection-ICDAR07}
Guangyu Zhu and David Doermann.
\newblock Automatic document logo detection.
\newblock In {\em Proceedings of the 9th International Conference on Document Analysis and Recognition (ICDAR)}, 2007.

\end{thebibliography}
}

\end{document}